%% file: root_arxiv.tex
\documentclass[letterpaper, 10 pt, conference]{ieeeconf}  

\IEEEoverridecommandlockouts                              

\usepackage{multicol}
\usepackage[bookmarks=true]{hyperref}
\usepackage[english]{babel}
\usepackage{amsmath,amssymb,amsfonts}
\usepackage{algorithm}
\usepackage{algpseudocode}
\usepackage{graphicx}
\usepackage{textcomp}
\usepackage{xcolor}
\usepackage{tabularx}
\usepackage{xspace}
\usepackage{url}
\usepackage{booktabs}
\usepackage{threeparttable}
\usepackage{array}
\usepackage{caption}
\usepackage{pifont}
\usepackage{bm}
\let\labelindent\relax
\usepackage{enumitem}
\usepackage{gensymb}

\definecolor{royalblue}{RGB}{65, 105, 225}
\definecolor{softgreen}{RGB}{85, 170, 85} 
\definecolor{softred}{RGB}{200, 50, 50}   

\hypersetup{
    colorlinks=true,
    linkcolor=orange,    
    citecolor=softgreen,    
    urlcolor=royalblue      
}

\newcommand{\system}{Locomotion Beyond Feet\xspace}

\newcommand{\largebox}{knee-high platforms}
\newcommand{\wall}{knee-high walls}
\newcommand{\stair}{steep ascending and descending stairs}
\newcommand{\chair}{low-clearance spaces under chairs}

\newcommand{\crawlchair}{{crawling under a chair}\xspace}
\newcommand{\climbontobox}{{climbing onto a platform}\xspace}
\newcommand{\rotateonbox}{{rotating on a platform}\xspace}
\newcommand{\climboffbox}{{climbing down from a platform}\xspace}
\newcommand{\climbwall}{{climbing over a wall}\xspace}
\newcommand{\climbupstairs}{{climbing upstairs}\xspace}
\newcommand{\climbdownstairs}{{climbing downstairs}\xspace}
\newcommand{\getup}{{getting up from crawling}\xspace}
\newcommand{\getdown}{{getting down to crawling}\xspace}
\newcommand{\getupprone}{{getting up from prone}\xspace}
\newcommand{\getupsupine}{{getting up from supine}\xspace}

\title{\LARGE \bf
Locomotion Beyond Feet
}

\author{
\authorblockN{ 
\textbf{Tae Hoon Yang\authorrefmark{1}} \quad
\textbf{Haochen Shi\authorrefmark{1}} \quad
\textbf{Jiacheng Hu\authorrefmark{1}} \quad
\textbf{Zhicong Zhang} \quad
\textbf{Daniel Jiang} \quad
\textbf{Weizhuo Wang} \\
\textbf{Yao He} \quad
\textbf{Zhen Wu} \quad
\textbf{Yuming Chen} \quad
\textbf{Yifan Hou} \quad
\textbf{Monroe Kennedy III} \quad
\textbf{Shuran Song\authorrefmark{2}} \quad \textbf{C. Karen Liu\authorrefmark{2}}
}
\vspace{1mm}
\authorblockA{
\authorrefmark{1}Equal contribution \quad
\authorrefmark{2}Equal advising
}
\vspace{1mm}
\authorblockA{Stanford University}
\vspace{1mm}
\authorblockA{
\textbf{\textcolor{magenta}{\url{https://locomotion-beyond-feet.github.io}}}
\vspace{-2mm}
}
}

\begin{document}

\thispagestyle{empty}
\pagestyle{empty}

\input{sections/00-abstract}
\input{sections/10-intro}
\input{sections/20-related_works}
\input{sections/30-method}
\input{sections/40-experiments}

\input{sections/50-conclusion}






\section*{ACKNOWLEDGMENT}
The authors would like to express their great gratitude to the helpful discussions from all members of Stanford TML and REALab.
This work was supported in part by the NSF Award \#2143601, \#2037101, \#2132519, \#2153854, Sloan Fellowship, and Stanford Institute for Human-Centered AI. 
The views and conclusions contained herein are those of the authors and should not be interpreted as necessarily representing the official policies, either expressed or implied, of the sponsors. 




\bibliographystyle{IEEEtran}
\bibliography{whole_body_loco}

\end{document}

%% file: sections/00-abstract.tex
\input{captions/f1-teaser}

\begin{abstract}

Most locomotion methods for humanoid robots focus on leg-based gaits, yet natural bipeds frequently rely on hands, knees, and elbows to establish additional contacts for stability and support in complex environments. This paper introduces Locomotion Beyond Feet, a comprehensive system for whole-body humanoid locomotion across extremely challenging terrains, including low-clearance spaces under chairs, knee-high walls, knee-high platforms, and steep ascending and descending stairs. Our approach addresses two key challenges: contact-rich motion planning and generalization across diverse terrains. To this end, we combine physics-grounded keyframe animation with reinforcement learning. Keyframes encode human knowledge of motor skills, are embodiment-specific, and can be readily validated in simulation or on hardware, while reinforcement learning transforms these references into robust, physically accurate motions. We further employ a hierarchical framework consisting of terrain-specific motion-tracking policies, failure recovery mechanisms, and a vision-based skill planner. Real-world experiments demonstrate that Locomotion Beyond Feet achieves robust whole-body locomotion and generalizes across obstacle sizes, obstacle instances, and terrain sequences.

\end{abstract}

%% file: captions/f1-teaser.tex

\twocolumn[{%
\renewcommand\twocolumn[1][]{#1}%
\maketitle
\includegraphics[width=\textwidth]{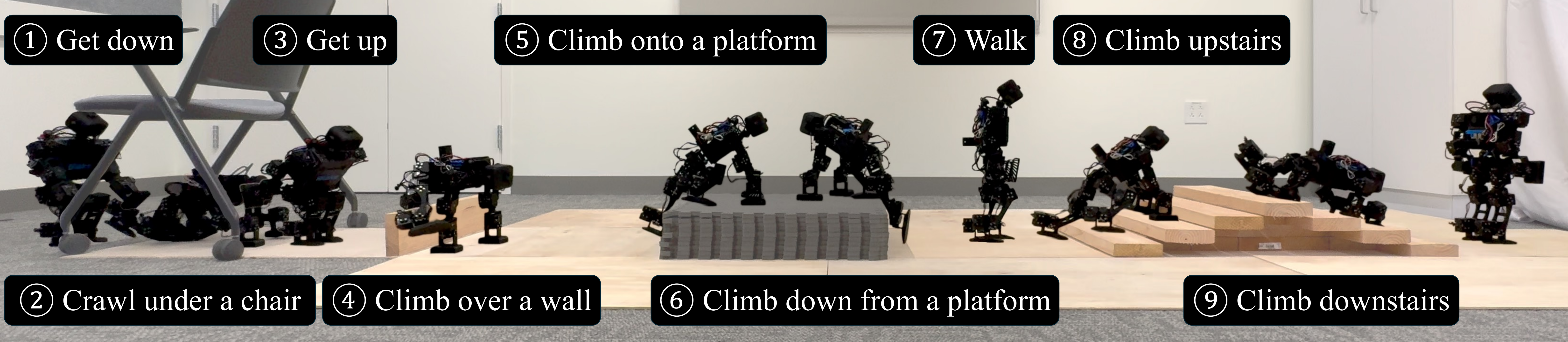}
\captionof{figure}{
\textbf{\system} enables whole-body humanoid locomotion on diverse and challenging terrains—including \chair, \wall, \largebox, and \stair—through chaining nine distinct locomotion skills that actively engage body parts beyond the legs for stability and support.}
\vspace{3mm}
\label{fig:teaser}
}]

%% file: sections/10-intro.tex
\section{Introduction}
\label{sec:intro}

Most locomotion methods for humanoid robots focus solely on leg-based movement~\cite{zhuang2024humanoida,allshire2025visual,hu2025robot}, yet bipeds in nature frequently leverage contacts from all limbs and torso to stabilize and support their bodies in complex environments~\cite{syeda2025phalangeal,mandery2015analyzing}. 
For example, in environments such as \chair, \largebox, \wall, and \stair, locomotion using only the feet becomes infeasible or necessitates abrupt motions. Humans naturally leverage additional body parts—such as hands, knees, and elbows—to establish extra contact points, enabling them to crawl, climb, and employ other whole-body strategies to overcome these obstacles.

We introduce a vision-based, hierarchical policy framework to enable highly diverse whole-body humanoid locomotion. Despite the benefits, whole-body locomotion remains underexplored in humanoid robots due to two main challenges: (1) Navigating complex environments requires strategic contact planning and robust control. (2) Different terrains require fundamentally different motor skills, such as walking, climbing, or crawling.

To address the first challenge, a key insight is that traditional keyframe animation and reinforcement learning (RL) are highly complementary for learning terrain traversal policies. Keyframe animation provides an intuitive approach to encode human knowledge of motor skills and physical interactions with the environment into robot control, such as specifying critical contact states and joint configurations~\cite{akgun2012trajectories}. Because natural human motion is typically low-frequency~\cite{khusainov2013realtime}, keyframes serve as an effective abstraction. 

Similar to prior approaches that retarget human motion capture (mocap) trajectories for motion tracking~\cite{fu2024humanplus,liao2025beyondmimic,chen2025handeye,qiu2025humanoid,ze2025twist,he2024omnih2o}, traditional keyframe animation provides kinematics but relies on RL trained in physics simulation to become dynamically viable robot policies. Importantly, unlike motion capture data, keyframes bypass the embodiment gap entirely by directly designing reference motions in the robot’s state space. This frees us from carefully matching human and robot embodiments, and instead allows exploration of the robot’s full hardware capabilities, producing motions not constrained by human demonstrations. Furthermore, the physical plausibility of keyframes can be verified in simulation and validated in the real world through open-loop execution, significantly accelerating the design iterations. In practice, once familiar with the tools, designing a physically consistent trajectory with keyframes typically requires only a few hours, even for challenging locomotion skills such as \climbwall, considerably more efficient than combined efforts of motion capture, human motion data retrargeting, and extensive reward shaping for RL training.

To address the second challenge that different terrain requires different skills, we argue that a single vision-based policy is not necessary and likely less desirable: a hierarchical framework is more robust. Our hierarchical framework allows diverse motion tracking policies tailored to distinct terrains, robust failure recovery mechanisms for fall situations, and a general vision-based planner that classifies terrain with stereo fisheye cameras and learned depth estimation. While rapid responses to local disturbances require fast 50 Hz control loops, locomotion mode selection with vision input can robustly operate at a lower frequency (3.1 Hz).

\input{captions/f2-pipeline}

As shown in Figure~\ref{fig:teaser}, \system is a comprehensive framework that enables traversal of extremely challenging obstacle courses through three categories of motor skills: (1) locomotion skills such as walking and crawling, (2) transition skills such as \getup, \getdown, \getupprone, and \getupsupine, and (3) terrain-specific skills such as \climbontobox, \rotateonbox, \climboffbox, \climbwall, \climbupstairs, and \climbdownstairs. Extensive real-world experiments demonstrate the system’s robustness to obstacle sizes, obstacle instances, and terrain sequences. All work is open-sourced.

%% file: captions/f2-pipeline.tex
\begin{figure*}[t]
  \centering
  \includegraphics[width=\textwidth]{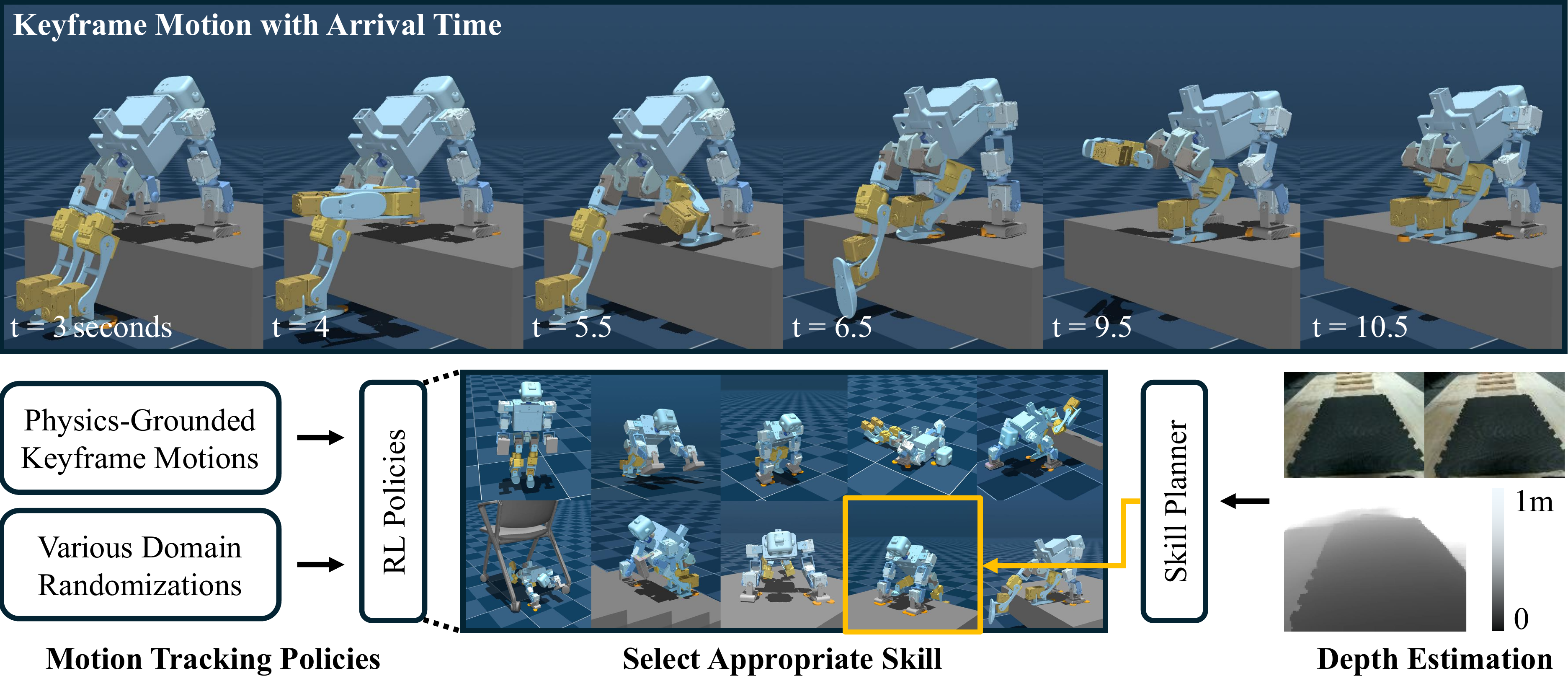}
  \vspace{-3mm}
  \caption{\textbf{System Pipeline.} First, we generate physics-grounded keyframe motions with a physics-aware GUI application, where robot poses and arrival times are specified interactively. Second, we interpolate the keyframes to create reference motions, which serve as tracking rewards for RL policies. We further apply extensive domain randomization, such as initial robot states, obstacle dimensions, frictions, and IMU noise. Finally, a skill planner processes depth input from a learned depth estimation module at $3.1~\mathrm{Hz}$, along with IMU readings and the current skill, to select the next  appropriate skill.}
  \label{fig:pipeline}
  \vspace{-3mm}
\end{figure*}

%% file: sections/20-related_works.tex
\section{Related Works}
\label{sec:related_works}

\subsection{Locomotion on Challenging Terrains}
Biomechanical studies reveal clear distinctions between quadrupedal and bipedal locomotion modes: macaques walking bipedally adopt a wider step width, longer duty cycle, and extended double-support phase to compensate for upright posture and a shifted center of mass~\cite{higurashi2019locomotor}. 
The key distinction lies in the size of the support polygon, defined as the convex hull of ground contact patches. Static balance is possible only when the center of mass (CoM) remains within this polygon. Leg-only locomotion yields a small support polygon that reduces to a single contact patch during footstep transitions, often necessitating abrupt motions on difficult terrain. In contrast, whole-body locomotion can employ three or more contact patches to form a larger and more consistent support polygon, yielding more stable and safer movement. 

On the robotics side, RL has enabled robust locomotion on challenging terrains; for instance, Rudin et al.~\cite{rudin2022learning} demonstrated massively parallel deep RL. However, the same strategy is applied to both quadrupedal and bipedal robots without accounting for their distinct locomotion modes.
Quadrupedal robots excel at terrain traversal through coordinated leg motion and have achieved agile parkour behaviors~\cite{zhuang2023robota,cheng2024extremea,rudin2025parkour,choi2023learning,lai2024world,hoeller2024anymal}.
Recent humanoid approaches~\cite{zhuang2024humanoida,allshire2025visual} largely adopt quadruped-inspired strategies, relying primarily on leg-based locomotion with minimal arm involvement and overlooking the distinct roles of arms and legs.
By contrast, our approach leverages whole-body motion, with all the body parts actively contributing to stability during terrain traversal, akin to natural human strategies in extreme environments.

\subsection{Keyframe Motion in Robotics}

Keyframes provide intuitive human control over motion synthesis, originating from character animation~\cite{lasseter1987principles,witkin1988spacetime,gleicher1997motion}. In robotics, keyframes have been adapted for humanoid motion generation through optimization~\cite{antonelli2009intuitive}, as reference trajectories for learning motion tracking policies~\cite{peng2018deepmimic,grandia2024design}, and as sparse rewards to achieve specific goals at predetermined times~\cite{zargarbashi2024robotkeyframing}. 
In these works, keyframes offer an intuitive mechanism for encoding human expertise and biomechanical insights into robotic motion synthesis~\cite{akgun2012trajectories}. The effectiveness of keyframe representation stems from the observation that natural human locomotion exhibits predominantly low-frequency characteristics~\cite{khusainov2013realtime}, making sparse temporal sampling through keyframes a well-suited abstraction that captures essential motion dynamics. Inspired by prior work, we leverage keyframe motions as references for training terrain-specific whole-body locomotion skills.

Unlike traditional keyframe animation, our approach ensures physics-grounded motions through a MuJoCo-integrated tool~\cite{todorov2012mujoco} that allows interactive visualization and validation of dynamics and contacts. Keyframing further distills human knowledge of dynamics into motion priors by explicitly specifying contact transitions and support phases. In contrast, recent kinematic retargeting methods~\cite{fu2024humanplus,liao2025beyondmimic,chen2025handeye,qiu2025humanoid,ze2025twist,he2024omnih2o} lack dynamics information~\cite{izani2003keyframe} and suffer from embodiment gaps. Moreover, while retargeting pipelines make RL training and sim-to-real transfer difficult to verify, keyframes can be validated directly in simulation or via open-loop execution on the real robot, enabling faster iteration and clearer optimization.

\subsection{Perception for Legged Locomotion.}
Legged locomotion has employed diverse perception modalities, using lidar point clouds for geometric understanding~\cite{hoeller2024anymal,ma2025ppl} and depth sensing for terrain perception~\cite{allshire2025visual,rudin2025parkour,lai2024world}. Our approach adopts depth-based perception for its accessibility and computational efficiency. Notably, advances in learned stereo models such as Foundation Stereo~\cite{wen2025foundationstereo} enable depth estimation directly from RGB inputs, even eliminating the need for depth sensors.

\subsection{Policy Chaining}

Policy chaining is typically achieved by treating motor controllers as modular skills and composing them through a high-level planner that determines when each controller is activated. The main challenge lies in the mismatch between the terminal state distribution of one policy and the start state distribution of the next. Prior methods address this either by carefully engineering compatible start and terminal states~\cite{shi2023robocook} or by leveraging deep learning frameworks to train composite behaviors with a meta-composer policy~\cite{lee2021adversarial,xu2023compositea,christmann2024expert}. In our work, we found that a carefully designed state machine, in which all policies are trained to start and end in one of four canonical poses—standing, crawling, lying prone, or lying supine—was sufficient to ensure smooth transitions between skills during execution.

%% file: sections/30-method.tex
\section{Method}
\label{sec:method}


\system enables whole-body locomotion through four key components: physics-grounded keyframe motion generation, DeepMimic-based motion tracking policies, a depth-conditioned visual skill classifier, and a hierarchical skill execution framework (Figure~\ref{fig:pipeline}).

\subsection{Physics-Grounded Keyframe Motion}

We generate reference motions using a GUI tool based on MuJoCo~\cite{todorov2012mujoco} that allows intuitive design of physically plausible motions. In the app, the user specifies robot poses along with their execution order and arrival times. The resulting keyframe sequence is then linearly interpolated to generate a complete trajectory. Although specifying arrival times may seem difficult, we found that simple choices such as $0.5$ seconds, $1$ second, or $2$ seconds are usually sufficient.

Keyframe motion is most criticized for the need for manual tuning~\cite{goel2025generative}. To mitigate this, we streamline keyframe design with utilities for joint mirroring, aligning the robot's feet to the ground, and visualization of the center of mass, collisions, and contacts. Our tool enables quick validation of individual keyframes and full trajectories for balance and smoothness. In practice, for simple motions such as crawling, we design the entire trajectory to be physically valid and directly replayable in simulation. For more challenging motions such as \climbwall, we instead ensure that individual keyframes are statically stable, so that the linearly interpolated trajectory remains physically plausible.

Another limitation of keyframe motion is its open-loop nature—it cannot adapt to perturbations, modeling errors, or unexpected environmental changes. While it provides a strong prior, it lacks the reactive flexibility required for real-world deployment. To address this, we train motion-tracking policies with RL that robustly execute keyframe motions while adapting to various uncertainties.

\subsection{Motion Tracking Policies}

We categorize motion tracking policies into three types:

\textbf{Locomotion skills} provide continuous control for periodic locomotion behaviors such as walking and crawling. These skills can be executed indefinitely and do not enforce a fixed terminal pose, making them suitable for continuous locomotion and seamless handoff to subsequent skills. In particular, the walking skill is modulated by velocity commands. We implement a command-conditioned walking policy that enables reactive control from high-level navigation inputs, such as walking forward and turning, while maintaining stable and robust locomotion.

\textbf{Transition skills} handle transitions between different poses, such as standing to crawling, crawling to standing, lying prone to standing, and lying supine to standing. 

\textbf{Terrain skills} handle specific terrains, including \climbontobox, \rotateonbox, \climboffbox, \climbwall, \climbupstairs, and \climbdownstairs. Each policy autonomously executes the entire trajectory once triggered.

We train all three types of skills following a similar recipe: RL-based motion tracking policies $\pi(\bm{\mathrm{a}}_t | \bm{\mathrm{s}}_t)$ that output joint position setpoints $\bm{\mathrm{a}}_t$ for proportional-derivative (PD) controllers. The observable state $\bm{\mathrm{s}}_t$ includes:
\begin{equation}
\bm{\mathrm{s}}_t = \left(\bm{\phi}_t, \bm{\mathrm{c}}_t, \Delta\bm{\mathrm{q}}_t, \bm{\dot{\mathrm{q}}}_t, \bm{\mathrm{a}}_{t-1}, \bm{\theta}_t, \bm{\omega}_t \right),
\end{equation}
where $\bm{\phi}_t$ denotes the phase signal for temporal coordination. The locomotion policies employ a periodic signal, whereas the transition and terrain policies use a monotonically increasing signal. $\bm{c}_t$ represents optional velocity commands, $\Delta\bm{q}_t$ is the joint position offset from the neutral pose $\bm{q}_0$, $\bm{\dot{q}}_t$ is the joint velocity, $\bm{a}_{t-1}$ is the previous action, $\bm{\theta}_t$ is the torso orientation, and $\bm{\omega}_t$ is the torso angular velocity.

\begin{algorithm}[t]
\caption{Skill Planner}
\label{alg:hierarchical}
\begin{algorithmic}[1]
\Require depth map $\bm{D}$; torso pitch and bounds $\theta, \theta_{\min},$ $\theta_{\max}$; depth extrema bounds $\delta_{\min}$, $\delta_{\max}$; confidence threshold $c$; current skill $S_{\mathrm{curr}}$; recovery skill $S_{\mathrm{rec}}$. 
\vspace{2pt}
\Function{SelectSkill}{$\bm{D}$, $S_{\mathrm{curr}}$, $\theta$}
    \State $\text{fallen} \gets 
    \big(\theta_{\mathrm{pitch}} < \theta_{\min} \;\textbf{and}\; \max(\bm D) < \delta_{\min}\big)$
\Statex \phantom{$\text{fallen} \gets$} $\;\textbf{or}\;
    \big(\theta_{\mathrm{pitch}} > \theta_{\max} \;\textbf{and}\; \min(\bm D) > \delta_{\max}\big)$
    \If{$\text{fallen}$}
        \State \Return $S_{\mathrm{rec}}$
    \EndIf
    \State $\bm{p} \gets \textsc{ClassifySkill}(\bm{D})$
    \State $\bar{\bm{p}} \gets 0.1\cdot\bar{\bm{p}} + 0.9\cdot\bm{p}$
    \State $S_{\mathrm{best}} \gets \arg\max_j \, \bar{\bm{p}}[j]$
    \If{$\bar{\bm{p}}[S_{\mathrm{best}}] > c$}
        \State \Return $S_{\mathrm{best}}$
    \Else
        \State \Return $S_{\mathrm{curr}}$
    \EndIf
\EndFunction
\end{algorithmic}
\end{algorithm}

\input{captions/f3-terrain}
\input{captions/f4-motion_tracking}
\input{captions/f5-depth}
\input{captions/f6-planner}
\input{captions/f7-robustness}

We train these policies using PPO~\cite{schulman2017proximal} with reward functions following standard practices~\cite{peng2018deepmimic}:
\begin{equation}
    \mathrm{r}_t = \mathrm{r}_t^{\text{imitation}} + \mathrm{r}_t^{\text{regularization}} + \mathrm{r}_t^{\text{survival}}.
\end{equation}

The imitation reward $\mathrm{r}_t^{\text{imitation}}$ enforces accurate tracking of reference motions generated from our keyframe interpolation. An exception is made for the walking skill, which does not use whole-body motion tracking. Instead, walking is trained using velocity tracking objectives together with a simple foot height tracking term that guides swing leg motion~\cite{zakka2025mujocoplayground, Shao_2022}, leading to an emergent walking pattern. Following prior work, additional stabilization rewards are included to guide torso orientation and foot posture during walking~\cite{Amazon_FAR_and_Abbeel_Holosoma}. The remaining exception is the rotating on a platform skill. While it retains whole-body motion tracking, it augments the imitation reward with a heading tracking term that encourages rotation about the vertical axis while remaining approximately in place. Although the reference motion corresponds to a crawling gait, this formulation enables the emergence of a stable in-place rotational behavior through coordinated whole-body contact. The regularization term $\mathrm{r}_t^{\text{regularization}}$ incorporates heuristics to minimize joint torques, energy consumption, and action rate, while the survival reward $\mathrm{r}_t^{\text{survival}}$ prevents early termination.

The term $\mathrm{r}_t^{\text{imitation}}$ is defined as a weighted sum across several tracking rewards. We follow similar formulation conventions in DeepMimic~\cite{peng2018deepmimic}: the pose reward $\mathrm{r}_t^p$ encourages alignment of body orientations with the reference motion, the velocity reward $\mathrm{r}_t^v$ matches local body velocities, the end-effector reward $\mathrm{r}_t^e$ tracks the positions of the hands and feet, and the center-of-mass reward $\mathrm{r}_t^c$ penalizes deviations of the robot’s center of mass from the reference trajectory.
\begin{equation}
    \mathrm{r}_t^{\text{imitation}} = \mathrm{w}_t^m\mathrm{r}_t^m + 
    \mathrm{w}_t^p\mathrm{r}_t^p +
    \mathrm{w}_t^v\mathrm{r}_t^v +
    \mathrm{w}_t^e\mathrm{r}_t^e +
    \mathrm{w}_t^c\mathrm{r}_t^c. 
\end{equation}
More specifically,$~\mathrm{r}_t^m$ is the motor position tracking reward:
\begin{equation}
\mathrm{r}_t^m \;=\; \sum_{g}
    \mathrm{w}_g \exp\!~\bigl(-\lVert \bm{\mathrm{q}}_g - \bm{\mathrm{q}}_g^{\ast} \rVert^2 \bigr),
\end{equation}
where $g \in \{\text{leg},\,\text{arm},\,\text{neck},\,\text{waist}\}$, $\bm{\mathrm{q}}_g$ is the actual motor position, $\bm{\mathrm{q}}_g^{\ast}$ is the reference motor position, and $\mathrm{w}_g$ is the weight assigned to action group $g$. For physically verified keyframe motion, we use the commanded actions as the motor position reference to account for tracking errors. 

To enable seamless sim-to-real transfer, we employ extensive domain randomization during training, including ground friction, motor actuation parameters, initial robot states, starting positions and orientations to different terrains, random pushes to robot bodies, IMU noise, and action delays. The IMU noise model combines colored noise, white noise, random-walk bias, and random amplitude scaling for $\bm{\theta}$ and $\bm{\omega}$, mimicking realistic IMU outputs.

All our polices are trained to start and end in either a standing pose, a crawling pose, a lying prone pose, or a lying supine pose. This design choice facilitates smooth transitions between different skills during execution.

\subsection{Visual Skill Classifier}

Our classifier enables autonomous skill selection by learning to classify appropriate skills from depth input.

\textbf{Data Collection.} To minimize the domain gap in depth observations, training data are collected directly from the real-world testbed. For each skill, RGB images are recorded from the robot’s dual-fisheye cameras at $30\,\mathrm{Hz}$, rectified using precomputed calibration parameters, and post-processed offline using Foundation Stereo~\cite{wen2025foundationstereo} to generate depth maps.

This offline processing is motivated by the fact that RGB capture runs at a significantly higher frame rate than depth estimation, allowing fast data collection without real-time depth inference. Obstacles are randomly positioned to create diverse terrains, and camera poses are slightly randomized within each rollout to increase visual diversity.

Locomotion skill data are collected throughout execution, while transition and terrain skill data are sampled only at the start of each skill. Ground-truth labels are assigned at data collection time, as transition moments between skills are well defined, temporally localized, and consistent across trials. As a result, collecting labeled training data incurs minimal overhead and does not require extensive manual annotation.

\textbf{Skill Classifier Training.} Skill classification is handled by a ResNet~\cite{he2016deep} classifier to select appropriate skills from depth input, striking a balance between computational efficiency and the ability to capture geometric cues such as obstacle shape, distance, and spatial layout. To address class imbalance, transition and terrain skills are weighted proportionally during training, as locomotion data are more abundant.

\textbf{Real-world Deployment.} During deployment, depth maps are generated from dual-fisheye RGB images using the same Foundation Stereo~\cite{wen2025foundationstereo} pipeline as in data collection, and are then fed into the skill classifier to select the skill label. The entire module runs at $3.1~\mathrm{Hz}$. 

\subsection{Hierarchical Policy Execution}

We introduce a hierarchical framework that separates vision-based planning from proprioception-based control, enabling modularity and robustness. The design employs a visual classifier running at $3.1~\mathrm{Hz}$, motivated by the observation that locomotion mode switching occurs at low frequency, coupled with a $50~\mathrm{Hz}$ low-level policy that enables rapid responses to local disturbances. 
Our framework also detects falls from IMU readings and triggers the recovery policies, further enhancing system robustness.

Our execution strategy is shown in Algorithm~\ref{alg:hierarchical}: during testing, depth maps are continuously processed and skills are predicted at $3.1~\mathrm{Hz}$. For smooth deployment, skill predictions are temporally stabilized with an exponential moving average. The system continues executing locomotion skills (walking or crawling) until the smoothed confidence surpasses a threshold, at which point it switches to the corresponding transition or terrain policy. Some skills are chained, such as \rotateonbox and \climboffbox. During a transition, classifier outputs are ignored; once it completes, the system resumes locomotion control until the next confident transition is triggered.

%% file: captions/f3-terrain.tex
\begin{figure}[t]
  \centering
  \includegraphics[width=\columnwidth]{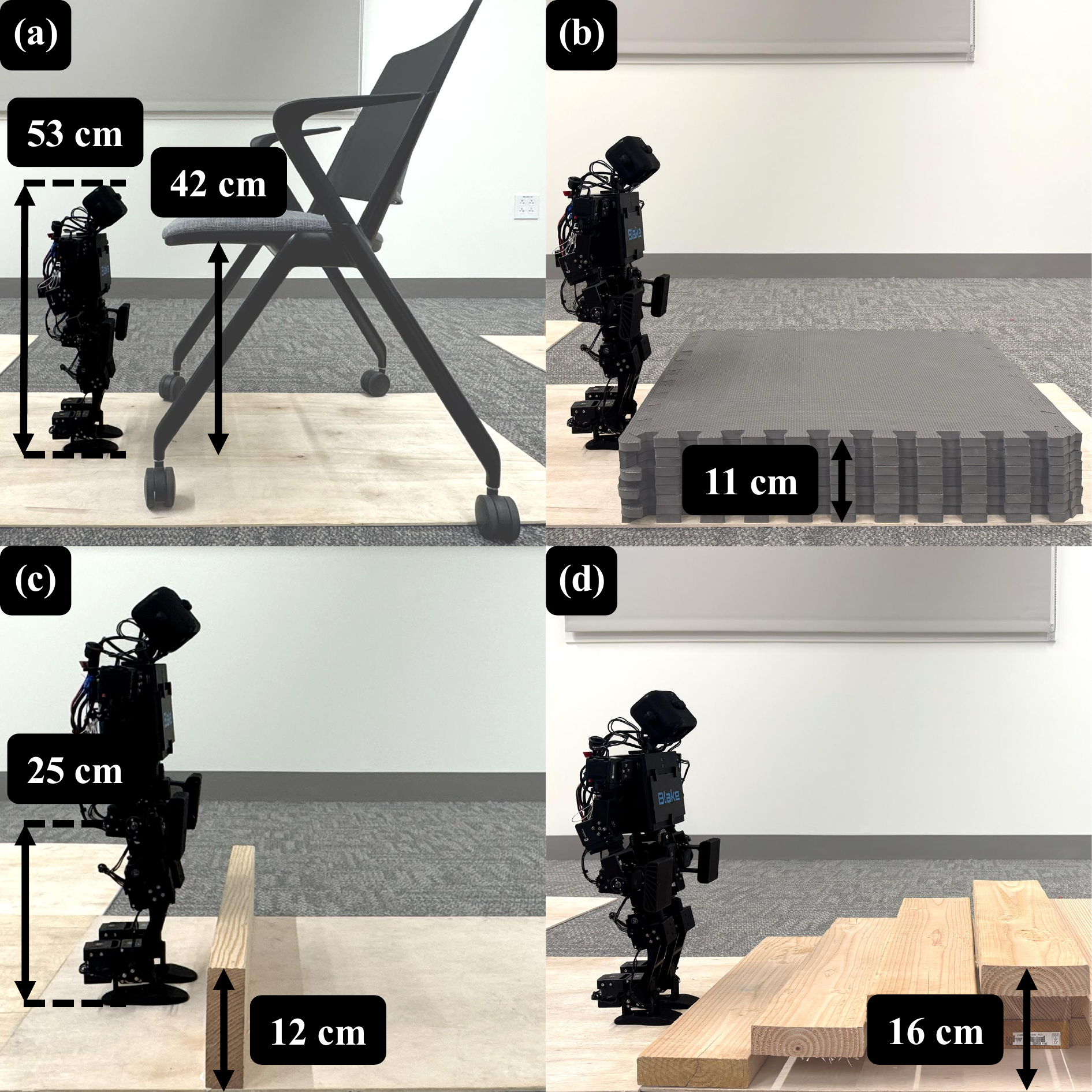}
  \caption{\textbf{Test Obstacles}. We show the robot beside test obstacles, including (a) \chair, (b) \largebox, (c) \wall, and (d) \stair. The space under the chairs is shorter than the robot ($53~\mathrm{cm}$), requiring crawling. The wall is 48\% of the robot's leg length ($25~\mathrm{cm}$), requiring climbing. The platform height is 44\% of the leg length, and each stair height is 16\% of the leg length, all posing extreme challenges at the robot’s scale.}
  \vspace{-3mm}
  \label{fig:terrain}
\end{figure}

%% file: captions/f4-motion_tracking.tex
\begin{figure*}[t]
  \centering
  \includegraphics[width=\textwidth]{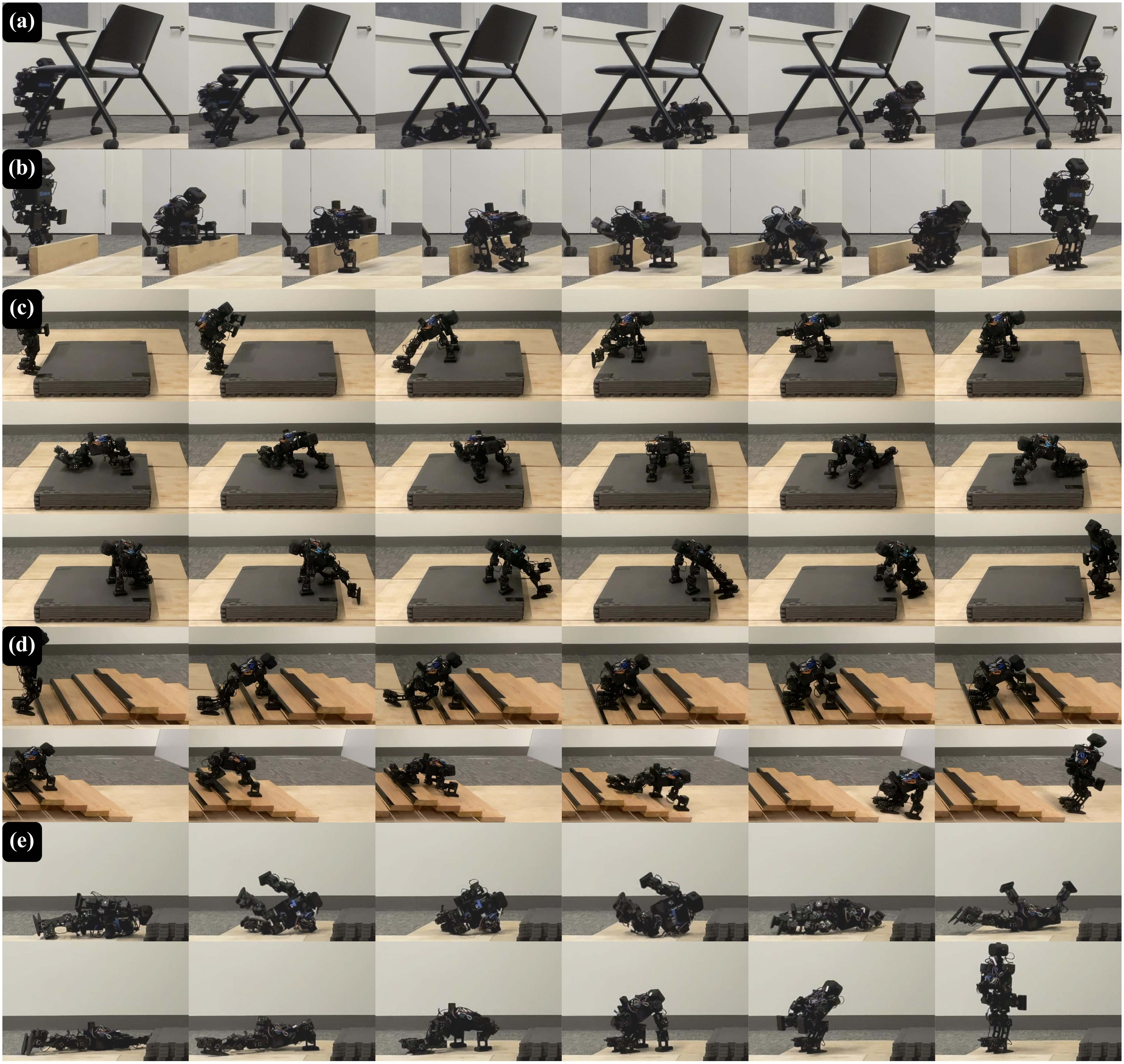}
  \caption{\textbf{Motion Tracking Policies.} We demonstrate our policies on traversing extremely challenging terrains—including (a) \chair, (b) \wall, (c) \largebox, and (d) \stair—and additionally show (e) fall recovery from supine and prone positions in case of failure.}
  \label{fig:motion_tracking}
  \vspace{-3mm}
\end{figure*}

%% file: captions/f5-depth.tex
\begin{figure}[t]
  \centering
  \includegraphics[width=\columnwidth]{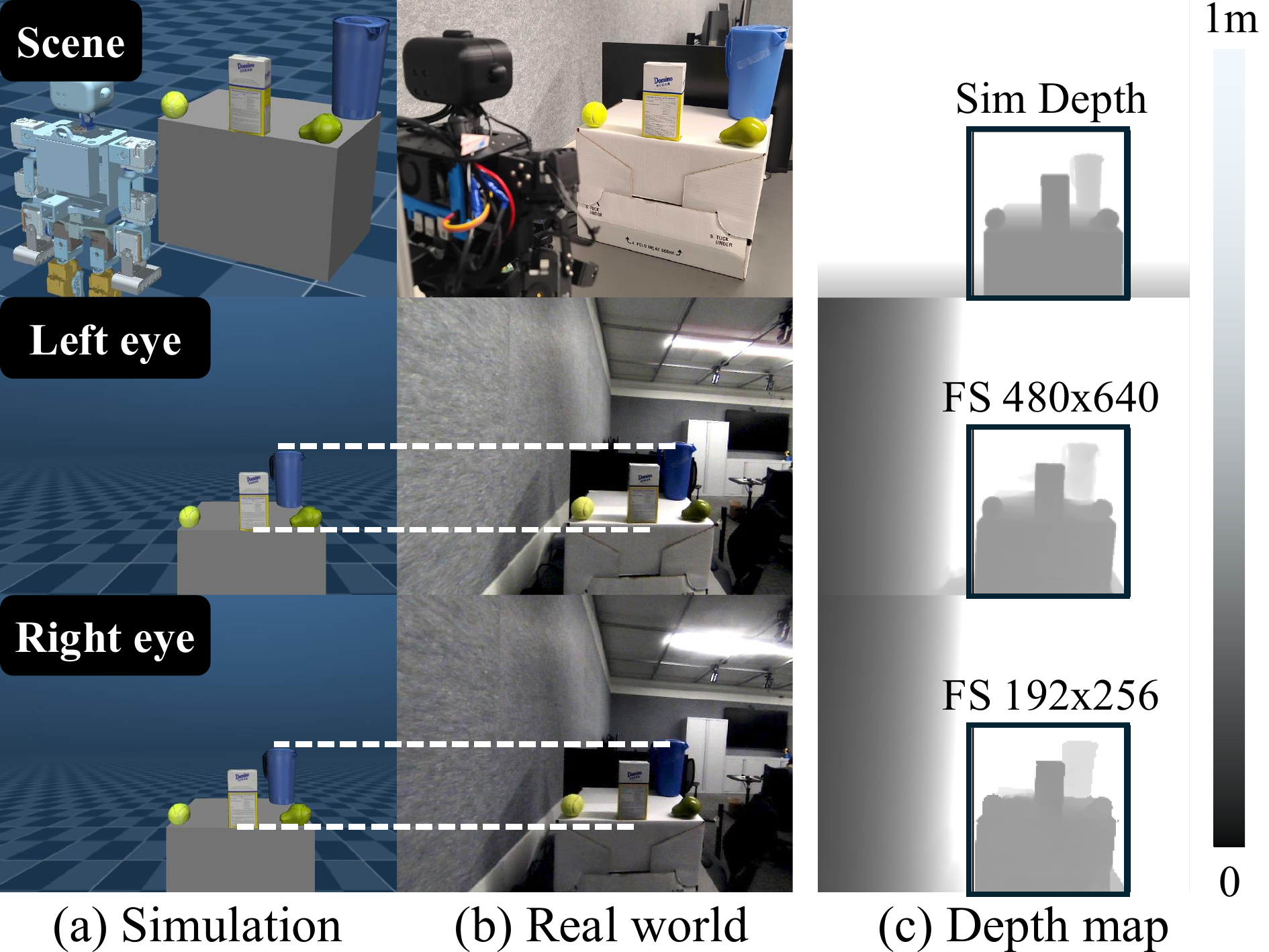}
  \caption{\textbf{Sim-to-real Depth Comparison}. We set up the same scene of YCB objects~\cite{calli2015ycb} (a) in MuJoCo~\cite{todorov2012mujoco} and (b) in the real world. The real-world RGB images are rectified after calibrating the fisheye cameras’ intrinsics and distortion, with white dashed lines illustrating proper alignment. (c) On the right is a comparison of ground-truth depth with real-world estimates from Foundation Stereo~\cite{wen2025foundationstereo} with resolution $480\times640$ and $192\times256$, respectively. We compute the quantitative results in the cropped region marked by the black box.} 
  \vspace{-3mm}
  \label{fig:depth}
\end{figure}

%% file: captions/f6-planner.tex
\begin{figure}[t]
  \centering
  \includegraphics[width=\columnwidth]{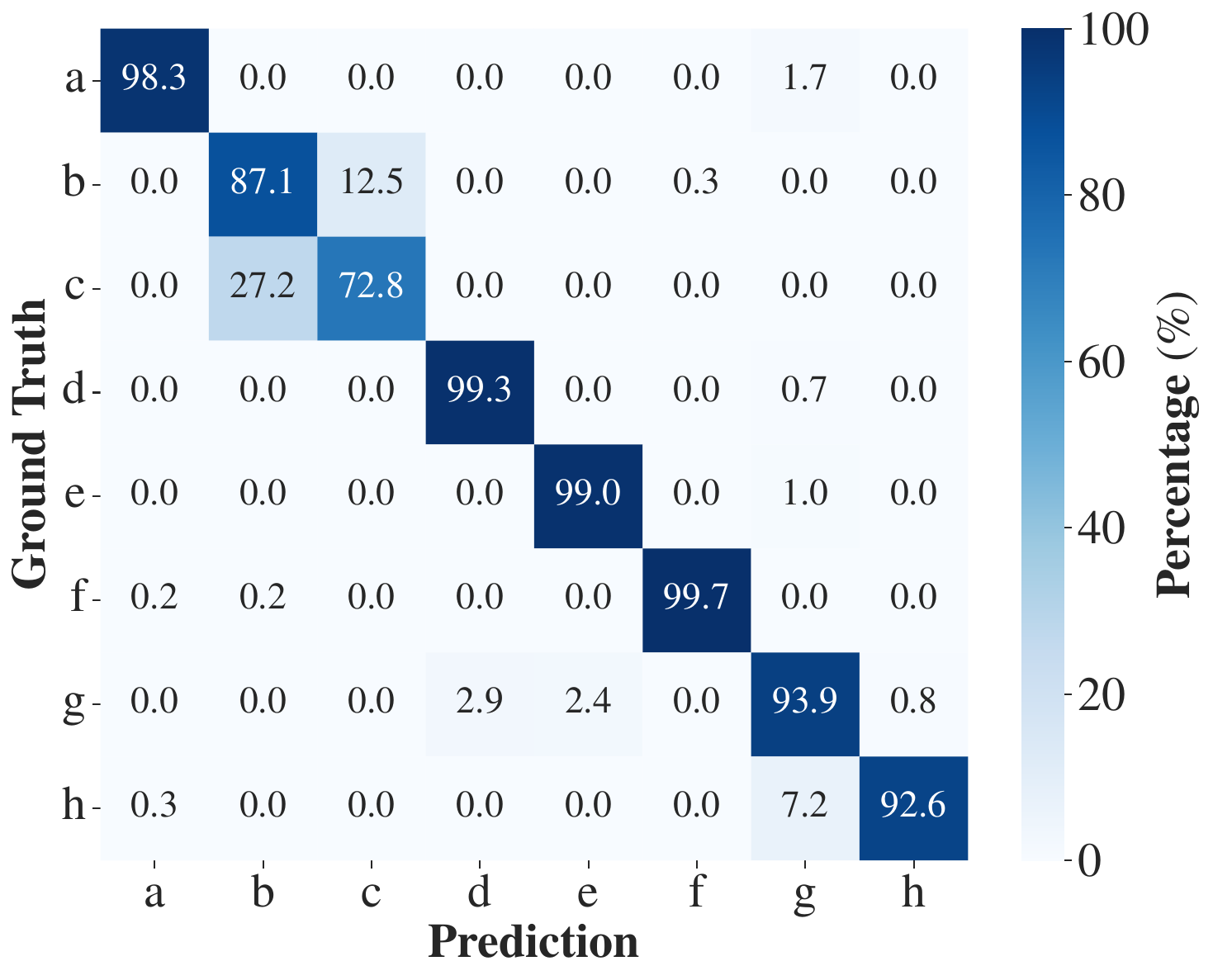}
  \caption{\textbf{Visual Skill Classifier Accuracy.} We present a confusion matrix of the skill classifier evaluated on the real-world test set. The skills are marked with letters: (a) \getdown, (b) \crawlchair, (c) \getup, (d) \climbwall, (e) \climbontobox, (f) \climboffbox, (g) walking, and (h) \climbupstairs.}
  \label{fig:planner}
  \vspace{-3mm}
\end{figure}

%% file: captions/f7-robustness.tex
\begin{figure}[t]
  \centering
  \includegraphics[width=\columnwidth]{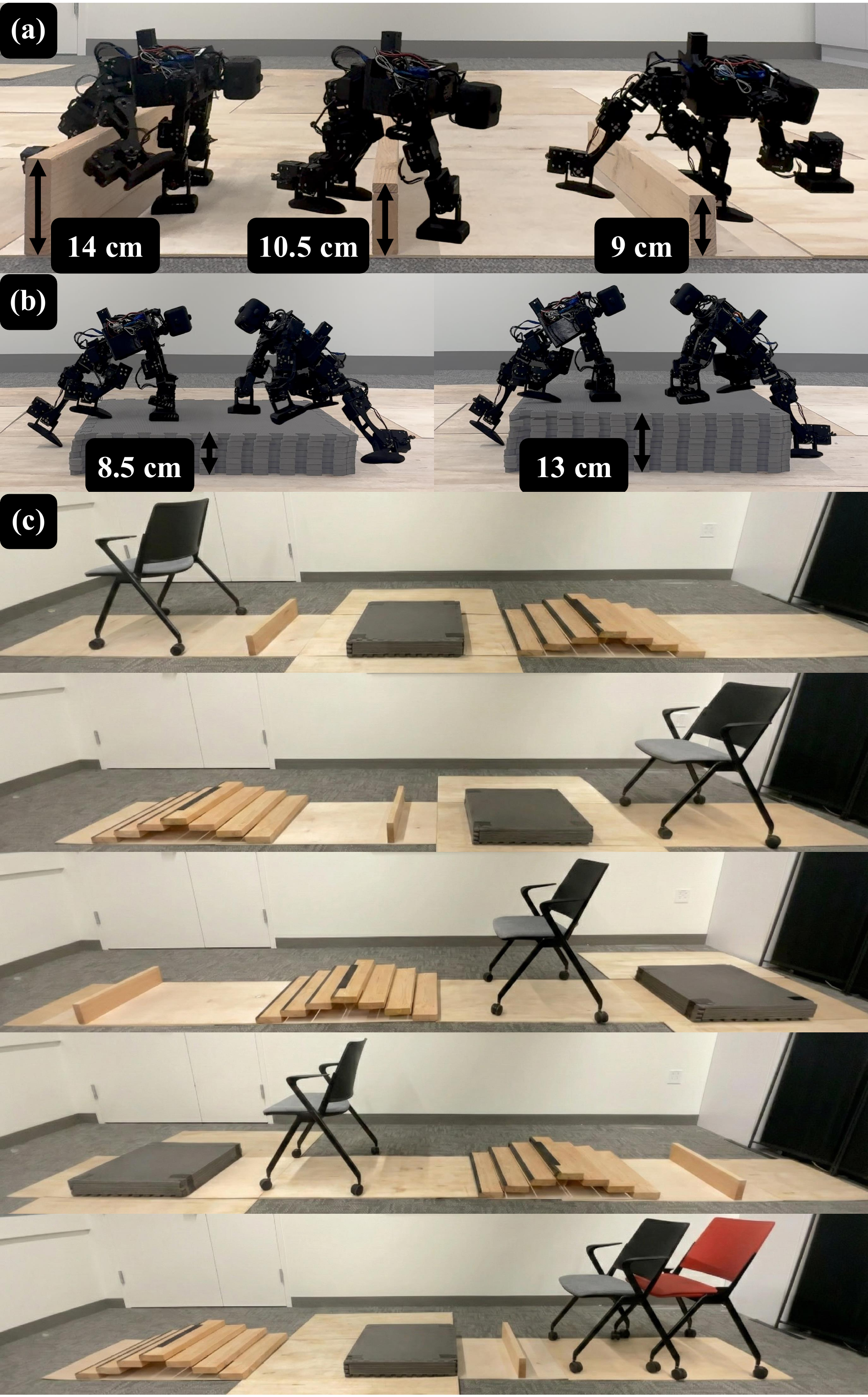}
  \caption{\textbf{System Robustness.} (a) The \climbwall policy designed for a $12~\mathrm{cm}$ wall generalizes to wall heights from $9~\mathrm{cm}$ to $14~\mathrm{cm}$. (b) Similarly, the \climbontobox policy designed for a $11~\mathrm{cm}$ platform generalizes between $8.5~\mathrm{cm}$ and $13~\mathrm{cm}$. (c) We demonstrate zero-shot success of our method across five obstacle orders and varying obstacle counts, such as two chairs in a row.}
  \label{fig:robustness}
  \vspace{-4mm}
\end{figure}

%% file: sections/40-experiments.tex
\section{Experiments}
\label{sec:experiments}

\subsection{Setup}

We use the open-source humanoid platform ToddlerBot~\cite{shi2025toddlerbot} for whole-body locomotion tasks. Its compact form factor, 30 degrees of freedom, and human-like range of motion make it well suited for evaluating terrain traversal skills in complex environments with diverse obstacles (Figure~\ref{fig:terrain}). These terrains are constructed using common household items such as chairs, foam blocks, and wooden planks, allowing for easy setup and reconfiguration.
To construct matching obstacles in simulation for policy training and evaluation, foam blocks and wooden planks are modeled with simple geometric primitives, while the chair’s geometry is captured using an image-to-mesh generator~\cite{chen2025ultra3d} with hand-measured scales for accurate representation.
We conduct controlled experiments on individual components as well as a holistic evaluation of the entire system.

\subsection{Motion Tracking Policies}

Figure~\ref{fig:motion_tracking} presents a selection of real-world photos, showcasing the effectiveness of our motion-tracking policies with zero-shot sim-to-real transfer, including \crawlchair, \climbwall, \climbontobox, \rotateonbox, \climboffbox, \climbupstairs, \climbdownstairs, \getupsupine, and \getupprone. The full experiment, including an additional cart-exit skill acquired with only a few hours of keyframe tuning, is shown in the supplementary video.

\subsection{Depth Estimation}

To improve the inference speed, we downsample the RGB resolution from $480 \times 640$ to $192 \times 256$ and compile the model using TensorRT with float16 precision, achieving $6\times$ speedup from $0.5~\mathrm{Hz}$ to $3.1~\mathrm{Hz}$ on a Jetson Orin NX 16GB without degrading skill classification accuracy.

To demonstrate the effectiveness of depth estimation with Foundation Stereo~\cite{wen2025foundationstereo}, we present a qualitative comparison in Figure~\ref{fig:depth}. Note that while some details are lost in the $192 \times 256$ version, they are unnecessary for terrain perception, as the obstacles are typically large structures. 
Quantitatively, our evaluation reports a pixel-wise mean absolute error (MAE) of $59~\mathrm{mm}$ within the black box region, as shown in Figure~\ref{fig:depth} and a point cloud Chamfer Distance of $17~\mathrm{mm}$, which primarily arises from misalignment between the simulation and the real-world scene setup. 
Moreover, the accuracy at $192\times256$ is comparable to $480\times640$, with a pixel-wise MAE of $62~\mathrm{mm}$ within the black box region in Figure~\ref{fig:depth} and a point cloud Chamfer Distance of $19~\mathrm{mm}$ compared to the ground truth in simulation. Empirically, the accuracy at $192\times256$ is sufficient for the downstream visual skill classifier.

\subsection{Visual Skill Classifier}

We train the skill classifier exclusively on real-world data, using a total of 27,987 depth map-skill pairs collected from the physical testbed, consisting of 22,389 training images and 5,598 validation images. Collecting the RGB images and corresponding skill labels for the training dataset required approximately $18$ minutes of real-world recording time, which is primarily determined by the fixed camera capture rate of $30\,\mathrm{Hz}$ rather than additional data collection overhead. Depth maps are generated offline by running the depth estimation model on the recorded RGB sequences, requiring approximately $150$ minutes of processing time. To evaluate generalization and robustness, we additionally collect 3,394 real-world samples as a held-out test set.

Figure~\ref{fig:planner} shows the confusion matrix evaluated on the real-world test set, achieving an overall classification accuracy of $93.9\%$. As shown in the confusion matrix, most misclassifications occur during transitional moments between skills, where geometric cues in the depth observations are inherently ambiguous. These cases often arise when obstacle distances or spatial layouts overlap across adjacent skills, such as between crawling under an obstacle and transitioning to a standing motion. Such misclassifications reflect natural ambiguity at skill boundaries rather than arbitrary prediction errors, and are effectively handled by the downstream skill planner through temporal smoothing and confidence-based selection (Algorithm~\ref{alg:hierarchical}).

\subsection{System Robustness}

We further evaluate the robustness of our policies by running them on terrains of varying scales, focusing on challenging skills, including \climbwall, \climbontobox, and \climboffbox. As shown in Figure~\ref{fig:robustness} (a) and (b), the terrain skills demonstrate strong robustness while relying solely on proprioceptive input, without visual information. Although the reference motions were designed for fixed terrain configurations (e.g., climbing over a $0.12~\mathrm{m}$ wall, and climbing onto and down from a $0.11~\mathrm{m}$ platform), policies trained with domain randomization on obstacle sizes generalize effectively to a wider range of obstacle heights. In contrast, simply replaying the keyframe animations and naive motion tracking policies will immediately fail when terrain configurations differ.
Moreover, we evaluate our hierarchical framework on five obstacle orders: (1) chair $\rightarrow$ wall $\rightarrow$ platform $\rightarrow$ stairs, (2) stairs $\rightarrow$ wall $\rightarrow$ platform $\rightarrow$ chair, (3) wall $\rightarrow$ stairs $\rightarrow$ chair $\rightarrow$ box, (4) platform $\rightarrow$ chair $\rightarrow$ stairs $\rightarrow$ wall, and (5) stairs $\rightarrow$ platform $\rightarrow$ wall $\rightarrow$ two chairs (Figure~\ref{fig:robustness}). All of these are successfully solved in a zero-shot manner as shown in the supplementary video.

%% file: sections/50-conclusion.tex
\section{Conclusion}
\label{sec:conclusion}

In conclusion, we present controlled experiments on individual components alongside a holistic system evaluation, demonstrating that \system achieves stable whole-body locomotion on challenging terrains—including \chair, \largebox, \wall, \stair—by actively engaging hands, knees, elbows, and other body parts to increase terrain contact. Although evaluated on a miniature humanoid, given the system robustness, we expect that our framework transfers seamlessly to full-size humanoids.


While our system achieves robust whole-body locomotion, several limitations remain. First, keyframe design requires manual effort and domain expertise, though this enables rapid iteration compared to human motion retargeting. Automated design through optimization could enhance scalability. Second, linear interpolation between keyframes trades motion naturalness for simplicity—more sophisticated interpolation methods could improve motion quality. Third, extreme contact-rich strategies like \climbwall that succeed in simulation occasionally fail on hardware due to contact modeling approximations. These limitations highlight interesting avenues for future research, while our current system provides a practical solution for diverse terrain traversal.